\documentclass[11pt,a4paper,hidelinks]{article}
\usepackage[hyperref]{emnlp2020}
\usepackage{times}
\usepackage{latexsym}
\usepackage{multirow}

\usepackage{subcaption}

\usepackage{microtype}

\aclfinalcopy % Uncomment this line for the final submission
\usepackage{times}

\usepackage{soul}
\usepackage{url}
\usepackage{hyperref}
\usepackage{amsmath}
\usepackage{booktabs}
\usepackage{times}
\usepackage{multicol}
\usepackage{lipsum}
\usepackage{mathtools,xparse}
\usepackage{float}
\usepackage{xspace}

\usepackage{cleveref}
\crefname{section}{\S}{\S\S}
\Crefname{section}{\S}{\S\S}
\crefname{table}{Table}{Tabs.}
\crefname{figure}{Figure}{}
\crefname{algorithm}{Alg.}{}
\crefname{equation}{Eq.}{}
\crefname{align}{Eq.}{}
\crefname{appendix}{Appendix}{}
\crefformat{section}{\S#2#1#3}  % remove space between section symbol and the number

\title{On the Complementary Nature of Knowledge Graph Embedding, Fine Grain Entity Types, and Language Modeling}

\author{
Rajat Patel \and
Francis Ferraro\\
University of Maryland, Baltimore County\\
{\tt\{rpatel12, ferraro\}@umbc.edu}
}

\date{}

\begin{document}

\maketitle

\begin{abstract}
We demonstrate the complementary natures of neural knowledge graph embedding, fine-grain entity type prediction, and neural language modeling. We show that a language model-inspired knowledge graph embedding approach yields both improved knowledge graph embeddings and fine-grain entity type representations. Our work also shows that jointly modeling both structured knowledge tuples and language improves both. %
\end{abstract}

\section{Introduction}

The surge in large knowledge graphs---e.g., Freebase \cite{Bollacker2008FreebaseAC}, DBpedia \cite{Auer2007DBpediaAN}, YAGO \cite{Suchanek2007YagoAC}---has induced knowledge graph-based applications. Properly making use of this structured knowledge is a prime challenge. Knowledge graph embedding [KGE]~\cite{Bordes2013TranslatingEF,Socher2013ReasoningWN} addresses this problem by representing the nodes (entities) and their edges (relations) in a continuous vector space. Learning these representations deduces new facts from and identifies dubious entries in the knowledge base. It also improves relation extraction~\cite{Weston2013ConnectingLA}, knowledge base completion~\cite{Bordes2013TranslatingEF} and entity resolution~\cite{Nickel2011ATM}.

Entity typing can provide crucial constraints and information on the knowledge contained in a KG. While historically this has been modeled as explicitly structured knowledge, and recent work has modeled the contextual language in order to make in-context entity type classifications, we argue that language modeling techniques provide an effective approach for modeling both the explicit and implicit constraints found in both structured resources and free-form contextual language. 

Meanwhile, while language modeling [LM] has historically been a core problem within natural language processing~\cite{rosenfeld1994phd}, recent deep learning advances have been very successful in convincing the community of the power and flexibility of language modeling~\cite[i.a.]{peters2018elmo,devlin-etal-2019-bert,yang2019xlnet}. %

\begin{figure}[t]
    \begin{center}
    \resizebox{.95\columnwidth}{!}{
    \includegraphics{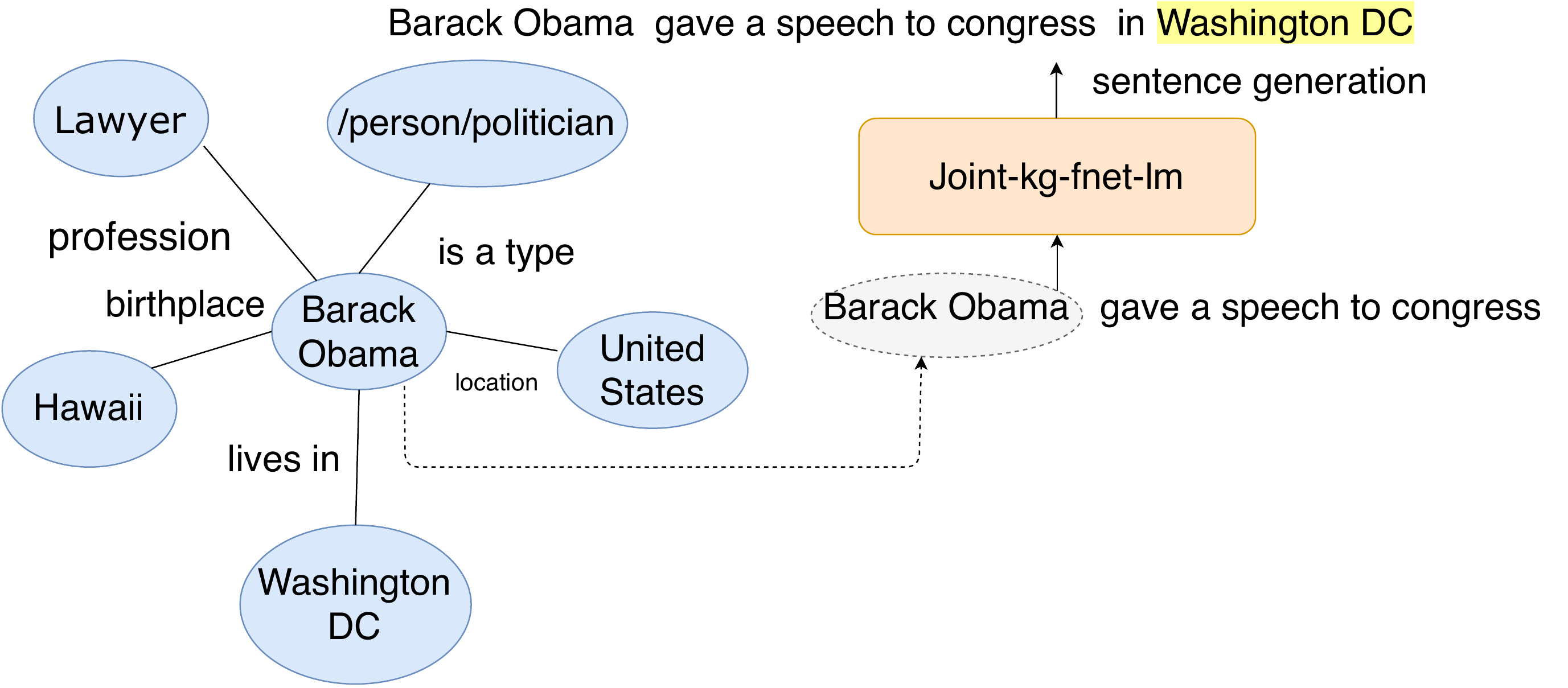}
    }
    \caption{Our joint learning framework learns the representation for the entity ``Barack Obama's'' in the same embedding space as that of the given input contextual description, ``Barack Obama gave a speech to Congress.'' %
    Further, by learning the entity type of `/person/politician', the model provides a better contextual understanding of the underlying entity.}
    \label{fig:jf_overview}
    \end{center}
\end{figure}

Building off of insights and advances in knowledge graph embedding, entity typing, and language modeling, we identify and advocate for leveraging the complementary nature of knowledge graphs, entity typing, and language modeling. %
In it, we introduce a comparatively simple framework that uses powerful, yet well-known, neural building blocks to (jointly) learn representations that simultaneously capture (1) explicit facts and information stored in a knowledge base, (2) explicit constraints on facts (exemplified by entity typing), and (3) \textit{implicit} knowledge and constraints communicated via natural language and discourse. \Cref{fig:jf_overview} provides an overview of the joint learning framework proposed in this work: an entity (``\textit{Barack Obama}'') along with its relations are represented in a continuous vector space. The framework also understands the underlying type (``\textit{/person/politician}'') for the given entity by learning the entity representation with contextual understanding (``\textit{Barack Obama gave a speech to Congress}''). By using the type and the factual information the framework enhances the comprehension of the focus entity in downstream applications like language modeling.\footnote{Though the entity typing examples here could be interpreted as being hierarchical, our method neither assumes nor requires any type hierarchy.} %

We note that others have explored what KG facts \textit{have already been learned} by \textit{specific}, advanced/contemporary LMs \cite{petroni-etal-2019-language}. %
That work utilized a pre-trained BERT model and queried what types of KG facts it contains. %
In addition, our primary goal is not broad, state-of-the-art performance---though we demonstrate that very strong performance is achievable. %
Rather, our goal is to examine what the complementary strengths, and evident limitations, of language modeling techniques for knowledge and entity type representation are. %
In doing so, we show that our joint framework yields empirical benefits for individual tasks. %
Our models leverage context-independent word embeddings, and we specifically eschew language models pre-trained on web-scale data.\footnote{
We do not deny that current pre-trained language models can be effective for other language-based tasks beyond language modeling. %
However, the reason we do not use transformer LMs like BERT or GPT-2 is because the amount of data they are pre-trained with can make it difficult to (a) fairly compare to previous work (is it the modeling approach, or the underlying, large-scale data at work?), and (b) identify and track the benefits of learning our tasks jointly.
} %
Our results further suggest that schema-free approaches to knowledge graph construction/embedding and fine grained entity typing should be studied in greater detail, and competitive, if not state-of-the-art, performance can be obtained with comparatively simpler, resource-starved language models. %
This has promising implications for low-resource, few-shot, and/or domain-specific information extraction needs.

Using publicly available data, our work has four main contributions. (1) It advocates for a  language-modeling based knowledge graph embedding architecture that achieves state-of-the-art performance on knowledge graph completion/fact prediction against comparable methods. (2) It introduces a neural-based technique based on both knowledge graph embedding and language modeling to predict fine-grain entity types, which yields competitive through state-of-the-art performance against comparable methods. (3) It proposes the joint learning of factual information with the underlying entity types in a shared embedding space. (4) It demonstrates that learning a knowledge graph embedding model and language model in a shared embedding space are symbiotic, yielding strong KGE performance and drastic perplexity improvements.\footnote{ %
Our code is available at \url{https://github.com/rajathpatel23/joint-kge-fnet-lm}.
}%

\section{Background}
The underlying information in the knowledge bases is difficult to comprehend and manipulate~\cite{Wang2014KnowledgeGE}. 
A vast number of knowledge graph embeddings techniques have been proposed over the years to mirror the entities and relations in the knowledge graphs. RESCAL~\cite{krompass2013non} is one of the first semantic-based embedding technique that captures the latent interaction between the entities and the relation. A model such as RESCAL can use graph properties to improve the underlying entity and relation representations \cite{Padia2019KnowledgeGF,balazevic2019tucker,Minervini2017RegularizingKG}. A more simplified approach is defined in DistMult ~\cite{Yang2014EmbeddingEA} by restricting the relation matrix to a diagonal matrix. 

Neural Tensor Network (NTN) ~\cite{Socher2013ReasoningWN} is one such technique that combines the relation specific tensors with head and tail vector representation over non-linear activation function mapped to hidden layer representation. Translational methods like TransE \cite{Bordes2013TranslatingEF} use distanced based models to represent entities and the relationships in the same vector space $R^d$. TransH \cite{Wang2014KnowledgeGE} overcomes the shortcomings of TransE by modeling the vector representation with relations specific hyperplane. TransR \cite{Lin2015LearningEA}, TransD \cite{Ji2015KnowledgeGE} model the representation similar to TransH by having relation specific spaces and decomposing the relation specific projection matrix as a product of two vector representations respectively.

Recognition of entity types into coarse grain types has been explored by researchers over the past two decades. Neural approaches have brought advances in extending the prediction problem from coarse grain entity types to fine-grain entity types. Work by \citet{Ling2012FineGrainedER}  was one of the first attempts in predicting the fine-grain entity types. The work framed the problem as multi-class multi-label classification. This work also led to an important contribution of a labeled dataset FIGER, widely used as a benchmark dataset in measuring the performance of fine-grain entity type prediction architectures. %
\citet{Ren2016AFETAF} introduced the method of automatic fine-grain entity typing by using hierarchical partial label embedding. %
\citet{Shimaoka2016AnAN} introduced a neural fine-grain entity type prediction architecture that uses semantic context with self-attention and handcrafted features to capture semantic context needed for fine-grain type prediction. \citet{Xin2018PutIB} showed that analyzing sentences with a pre-trained language model enhanced prediction performance. \citet{Zhang2018FinegrainedET} introduced a document level context and signifies the importance of mention level attention mechanism along with the sentence-level context in enhancing the performance of fine-grain entity prediction. \citet{Xu2018NeuralFE} enhanced neural fine-grain entity typing by penalizing the cross-entropy loss with hierarchical context loss for the fine-grain type prediction.

Language modeling has seen great progress in recent times. \Citet{Bengio2000ANP} pioneered the renewed use of distributed representation for dealing with the dimensionality curse imposed by the statistical methods. Their language model used recurrent neural networks for dealing with long sequences of text.  \citet{Mikolov2010RecurrentNN} extended the idea of building the recurrent neural network-based language models with an improved feedback mechanism of backpropagation in time. 

We are not the first to examine the intersection of knowledge graph embedding and language modeling. \Citet{ristoski2016rdf2vec,cochez2017global} directly embed RDF graphs using language-modeling based techniques. %
\Citet{Ahn2017ANK} and \citet{RobertLLogan2019BaracksWH} have more recently leveraged information from a knowledge base to improve language modeling. However, in addition to knowledge graphs and language modeling, we additionally consider fine-grain entity typing. %

With the success of contextualized vector representations and the availability of large-scale, pre-trained language models, there have been a number of efforts aimed at improving the knowledge implicitly contained in word and sentence representations. For example, \citet{bosselut-etal-2019-comet} introduce COMET, which describes a framework to learn and generate rich and diverse common-sense descriptions via language models (e.g., the autoregressive GPT-2). Similarly, \citet{zhang-etal-2019-ernie} and \citet{peters-etal-2019-knowledge} provide insights into aspects of LM on downstream NLP tasks. While we share the overall goal of improving knowledge representation within language modeling, the short-term goals are different, as we focus on individual facts, rather than traditional background/commonsense knowledge, and demonstrating the complementary nature of KGE, entity typing, and LM. %

\renewcommand{\baselinestretch}{1}
\begin{figure}[t]
\centering
    \includegraphics[scale=.25]{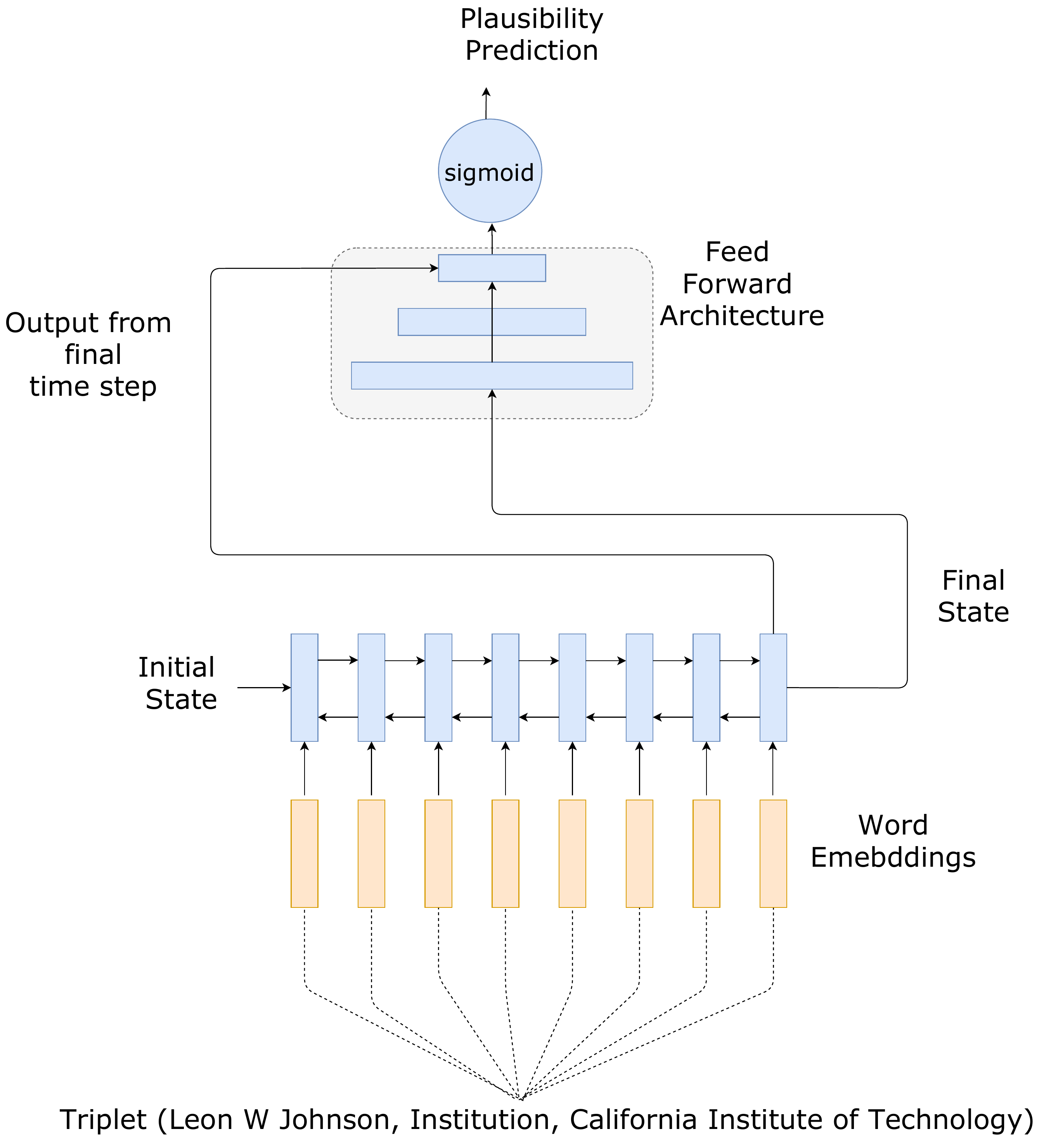}
    \caption{Knowledge Graph Embedding as language modeling, where triples are ``tokenized'' into word embeddings and the computed, sequential output states are used to predict triple correctness.}
    \label{fig:KGE_1}
\end{figure}

\section{Methodology}
This section introduces the framework for jointly learning knowledge graph embedding (KGE), fine grain entity types (ET) and language models (LM). It uses a multi-task learning architecture built over baseline architectures for all three tasks. We begin by introducing LM-inspired knowledge graph embedding and fine grain entity typing architectures; we describe the joint learning architectures in \cref{sec:results}. Fundamentally, our approach relies on appropriate and select parameter sharing across the KGE, ET, and LM tasks in order to learn these models jointly. While joint learning or multi-task learning through shared parameters have been examined before for a number of tasks, we argue that this parameter sharing is a very effective way to improve KGE, ET, and/or LM (for a particular baseline). Its simplicity is a core benefit.

\subsection{Knowledge Graph Embedding as a Language Model}
\label{sec:kge-as-lm}
The architecture in \cref{fig:KGE_1} embeds the factual entities and the relations. %
Let $G$ be a knowledge graph (KG) with nodes $V$ and edge $E$, where $V$ is a set of entities ${e_1, \dots, e_{|V|}}$ which are connected to each other by edges $E$. %
$E$ is a set of $K$ relations ${r_1, \ldots, r_k}$.
The architecture learns to embed the entities and relations into a (traditionally dense) vector space.
Given the head entity $e_i$, relation $r_k$ and tail entity $e_j$, we predict whether a given triplet $x_i = (e_i, r_k, e_j)$ is true (in the KG).

The model is a combination of a bi-LSTM~\cite{Hochreiter1997LongSM,Schuster1997BidirectionalRN} and a feed-forward architecture. In the spirit of language modeling, we represent each triple $x_i$ input to the architecture as a sequence of $n$ tokens $(x_{i_1}, x_{i_2}, .., x_{i_n})$. These tokens are represented in a continuous vector space by vector $v_{i_t}$ with dimension $d$, where $v_{i_d} \in {R^d}$. The bi-LSTM layer produces a learned representation of each token by maintaining two hidden states for each word: the forward state $\overrightarrow{h_{i_t}}$ learns representation from left to right (\cref{eq:1}) and the backward state $\overleftarrow{h_{i_t}}$ learns the representation from right to left (\cref{eq:2}):
\begin{align}
        &\overrightarrow{h_{i_t}} = \textrm{bi-LSTM}(W_{\overrightarrow{h}}x_{i_t} + V_{\overrightarrow{h}} \overrightarrow{h}_{i_{t-1}} + b_{\overrightarrow{h}})  \label{eq:1} \\
        &\overleftarrow{h_{i_t}} = \textrm{bi-LSTM}(W_{\overleftarrow{h}}x_{i_t} + V_{\overleftarrow{h}}\overleftarrow{h}_{i_{t+1}} + b_{\overleftarrow{h}}) \label{eq:2} \\
    &h_{i_t} = \mathrm{concat}[\overrightarrow{h_{i_t}}, \overleftarrow{h_{i_t}}]. \label{eq:3}
\end{align}
The forward and the backward states of the bi-LSTM layer are concatenated to produce a sequentially encoded representation $h_i$ for each time step $t$ given the input sequence $x_i$. %
The bi-LSTM weight matrices $W$ and $V$ and $b$ are learned during training. In principle the bi-LSTMs can be stacked, though we found not stacking to be empirically effective.

Though the bi-LSTM produces a sequence of hidden states, we summarize the information captured by it in a single, ``final'' state $C_{\textrm{final}}$. This state is then used to represent the information encoded by the whole sequence for the subsequent classification task. We let the rightmost state represent the ``final'' state, i.e.,  $C_{\textrm{final}} = 
h_{i_n}.$\footnote{In early experiments we tried other approaches, such as averaging all hidden representation to compute the final state ($C_{\textrm{final}} = \frac{1}{n}\sum_t h_{i_t}$. These caused neither large improvements nor decreases in performance. As a result, we advocate here for the simpler computation of $C_{\textrm{final}} = h_{i_n}$.}

The feed-forward architecture is a multi-layer perceptron with $L=3$ rectified linear hidden layers (ReLU). The input to the feed-forward layer is a  learned final cell state representation $C_{\textrm{final}}$ from the bi-LSTM sequence encoder. The feed-forward process captures the information from the learned sequence encoder and outputs a transformed representation $z_{l}$ from the final output layer: $z_l = \mathrm{ReLU}(W_{l} z_{l-1} + b_{l})$, with $z_{0} = C_{\textrm{final}}$, and layer-specific weights $W_{l}$ and biases $b_{l}$.

The output representation $z_{L}$ is then used to calculate the semantic matching score for the factual input $x_t$. This score is calculated by incorporating the learned representation $z_L$ with the sequentially encoded final sequence step  representation 
$h_{\textrm{t}}$. The product is then passed through a sigmoid activation function, $f(x_t; \theta) = \sigma(z_{l}^T h_{t}),$ where $\theta$ is a collection of network parameters used for training the language model-inspired knowledge graph embedding architecture. These parameters are jointly learned by minimizing a weighted cross-entropy loss with $\ell_2$ regularization (\cref{eq:9}):
\begin{multline}
    J(\theta) = -\frac{1}{N} \sum_{i=0}^{N} k \cdot y_i \cdot \log(f(x_t; \theta)) + \\ (1 - y_i) \cdot \log(1 - f(x_t; \theta)) + \lambda ||\theta||^2 \label{eq:9}
\end{multline}
where $k$ is the weight assigned to the positive samples during the training,  $y_i$ represents the original labels, and $\lambda$ is the regularization parameter.

As a result of our KGE method, we do not produce or store single, canonical representations of entities and relations. We argue that the lack of a canonical entity embedding is a large benefit of our model. First, it is consistent with the push for contextualized embeddings. Second, we believe that, even in a KG, an entity’s precise meaning or representation should depend on the fact/tuple that is being considered.\footnote{If a single representation is needed, note that because we tokenize entity, types, relations, and arguments into words, we could generate a single representation by combining the, e.g., entity’s individual word embeddings according to the LM.}

\subsection{Neural-Fine Grain Entity Type Prediction}

\begin{figure}[t]
    \begin{center}
    \includegraphics[width=\linewidth]{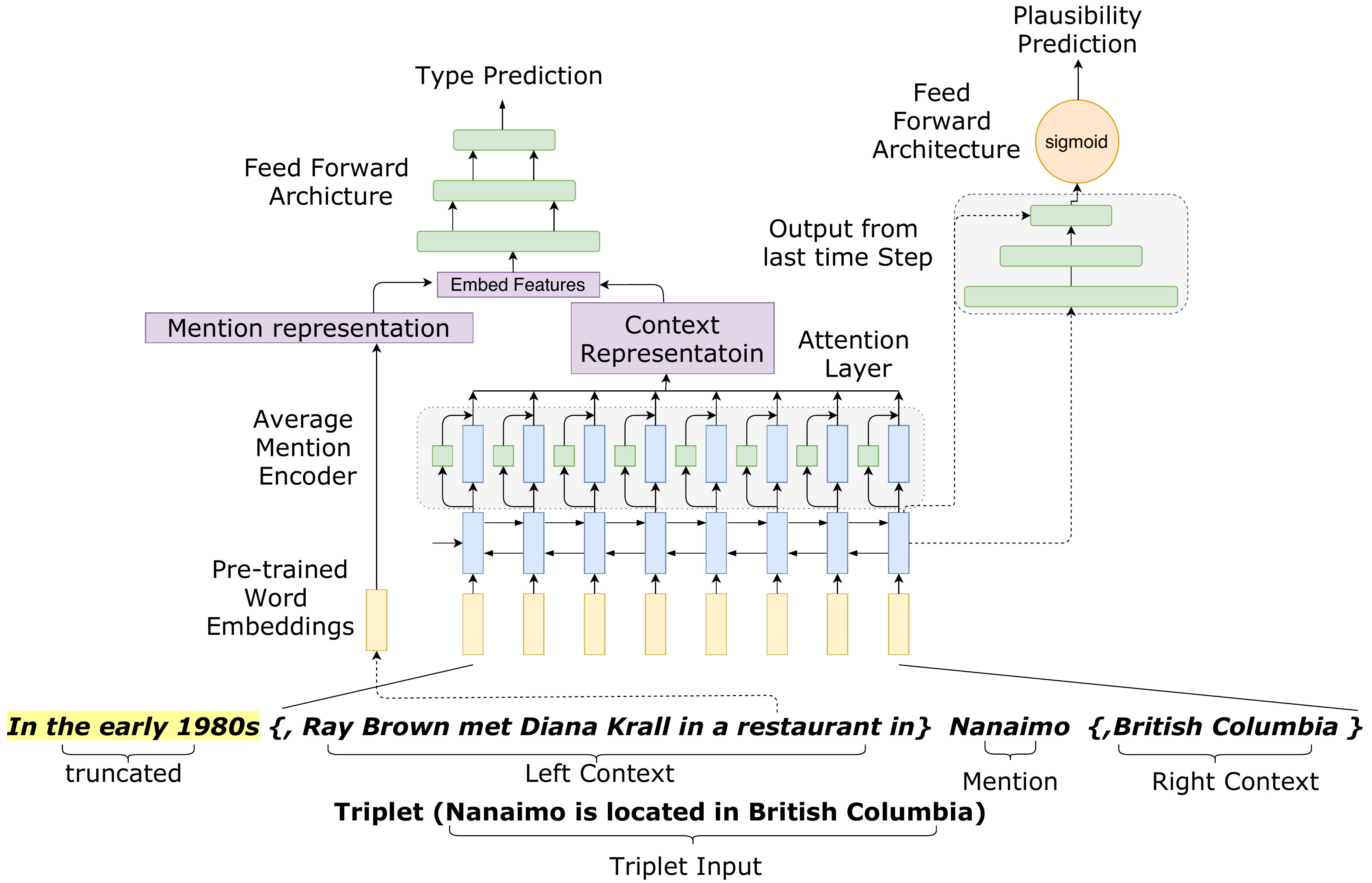}
    \caption{The joint learning architecture for training KGE and entity typing takes in both factual triplets and context information for an entity. Parameters of the architecture are trained to learn both the factual information as well the corresponding entity types.}
    \label{fig:KGE_FNER}
    \end{center}
\end{figure}

Recognizing the type of the given entity has been an integral part of tasks like knowledge base completion \cite{Bordes2013TranslatingEF}, question answering and co-reference resolution. \Citet{Ling2012FineGrainedER} extended the problem of entity type prediction to fine-grain entity types. 
Given an input vector $V_x$ for entity $x$, type embedding matrix $\theta$,  the function $g$ predicts all the possible entity types $t$ for given entity $x$ as $g(V_x; \theta) = \theta^T V_x$. The model learns the parameters $\theta$ by optimizing the hinge loss to classify a given entity into all the possible types T:
\begin{equation}
     J(\theta) = \sum_{t=0}^T \max(0, 1 - y_t \cdot g(V_x;\theta)).
\end{equation}
An entity is predicted to be of type $t$ if $g(x; \theta)$ is greater than a given threshold value $\tau$ (typically, $\tau = 0.5$, though it can be set empirically).

The architecture in \cref{fig:KGE_FNER} shows different sets of embedding-based features used to predict the entity type $t$. Word-level features and context level features---word spans to the left and right of the entity---are taken into consideration. The feature design used here is similar to the design of the features introduced by \citet{Shimaoka2016AnAN}. %
We note that our method neither assumes nor requires any type hierarchy, though including a type hierarchy is an avenue for future exploration.

\paragraph{Mention Encoder}
We encode a mention representation $m$ as the average of word embedding vectors $u_i$ for all words $i$ present in the given entity $e$: $m = \frac{1}{|n|} \sum_{i=0}^{n} u_i.$

\paragraph{Context Encoder}
The contextual representation for the given mention $e$ is performed by dividing into left context $l_c$ and right context $r_c$, where the left context is all the words present on the left of the given entity $e$, and the right context contains all the words present to the right of the given entity $e$. %
The left and right context are encoded by passing the context through a bi-LSTM sequence encoder \cite{Hochreiter1997LongSM,Schuster1997BidirectionalRN}. The sequence encoder is similar to the one used by \citet{Zhang2018FinegrainedET}. The outputs of the bi-LSTM sequence encoder are the sequential vector representation from both forward (left-to-right) and backward pass (right-to-left), $(l_f, l_b) = \textrm{BiLSTM}(l_c,h,h_{t-1})$ and $(r_f, r_b) = \textrm{BiLSTM}(r_c, h, h_{t-1}),$ where $(l_f, l_b)$ are the sequential output for the left context from forward and backward passes, $(r_f, r_b)$ are the sequential outputs from the right context from forward and backward passes, $h$ and $h_{t-1}$ are the current and the previous hidden states for forward and backward passes respectively. Left outputs are concatenated to form a left-looking encoding $L_c = \mathrm{concat}[l_f, l_b]$, while right outputs are concatenated to form a right-looking encoding $R_c = \mathrm{concat}[r_f, r_b]$. The complete contextual representation $C$ of the context is the concatenation of the left context and right context representations, $C =  \mathrm{concat}[L_c, R_c]$.

\paragraph{Attention}
We use an attention mechanism to reweight contextualized token embeddings. The attention layer, similar to that of \citet{Shimaoka2016AnAN}, is a 2 layer feed forward neural architecture where the attention weight for each time step of the context representation is learned given the parameter matrix $W_a$ and $W_s$: $a_i = \mathrm{softmax}(W_s \tanh({C_i \cdot W_a}))$. %
The context representation is a weighted sum of attention and the context representation,
$C_{rep} = \sum_{i=0}^{t} a_i \cdot C_i.$

The attention mechanism used here differs from \citet{Shimaoka2016AnAN} such that in our work the contextual embeddings share the same attention parameters. The features extracted from the mention encoder $m$ and attention weighted context encoder $C_r$ are concatenated to form a learned representation $V = \mathrm{concat}(m_i, C_{rep})$ that is passed to the feed-forward architecture for classification.

The feed-forward architecture is a 3-layer neural architecture with a batch normalization layer \cite{Ioffe2015BatchNA} present between the first and the second layers with a ReLU activation~\cite{Nair2010RectifiedLU}. The input to the feed-forward layer is a concatenated representation from the context and mention encoders. The feed-forward process captures the information from the learned features and outputs a transformed representation $q_{l} = \max(0, V_{l} \cdot q_{l-1} + d_{l})$ from the final output layer to classify the given mention into the corresponding entity types, where $V_{l}$, $d_{l}$ are the weights and bias for the hidden layer unit $l$ respectively. %$q_{l}$ represent the activation output from the hidden unit $l$. 
We initialize $q_0 = C_r$.

\subsection{Language Model}
The language model predicts the next possible word based on the previous inputs, as $p(w_n|w_1, w_2, ...w_{n-1})=\prod_i P(w_n| w_{n-k}, ....w_{n-1}).$ We use a simple LSTM to learn the sequential structure of the text.

\section{Experimental Settings}
The input to the joint learning architectures are the pre-trained GloVe embedding vectors trained on 840 billion words ~\cite{Pennington2014GloveGV}. The parameters of the baseline and the joint learning architecture are learned with Stochastic Gradient Descent and Adam ~\cite{Kingma2014AdamAM} as a learning rate optimizer. %
The training of the joint learning networks is performed with alternating optimization. The loss functions of the respective tasks are optimized at each alternate epoch/ interval. The hyper-parameters for training these joint architecture are chosen manually for the best-performing models on validation sets.

\paragraph{Data}
For a direct comparison of the performance as possible, we use previously studied datasets. We evaluate KG triple classification using the standard datasets of WordNet 11 (WN11) and Freebase 13 (FB13). WN11 \cite{Strapparava2004WordNetAA} is a publicly available lexical graph of \textit{synsets} (synonyms). Freebase \cite{Bollacker2008FreebaseAC} is a collaborative ontology  consisting of factual tuples  of entities related to each other through semantic relation. %
While recent work has advocated for examining variants and other derivatives of these datasets such as FB15k-237 and WN18RR~\cite[i.a.]{toutanova2015observed,dettmers2017convolutional,Padia2019KnowledgeGF}, there is a relative lack of previous experimental work on these newer datasets. Given space limitations, and in order to compare to the vast majority of previous work, we chose to report on the more common WN11 and FB13.

\begin{table}[t]
\begin{center}
\resizebox{.98\columnwidth}{!}{
 \begin{tabular}{llll}
\toprule
Method & WN11 & FB13 & Avg \\ \midrule
NTN \cite{Socher2013ReasoningWN} & 86.2 & 90.0 & 88.1 \\
TransE \cite{Bordes2013TranslatingEF} & 75.9 & 81.5 & 78.7 \\
TransH \cite{Wang2014KnowledgeGE} & 78.8 & 83.3 & 81.1 \\
TransR \cite{Lin2015LearningEA} & 85.9 & 82.5 & 84.2 \\
TransD \cite{Ji2015KnowledgeGE}  & 86.4 & 89.1 & 87.8 \\
TEKE \cite{Wang2016TextEnhancedRL}& 86.1 & 84.2 & 85.2 \\
TransG \cite{Xiao2016TransGA}  & 87.4 & 87.3 & 87.4 \\
TranSparse \cite{Ji2016KnowledgeGC}  & 86.4 & 88.2 & 87.4 \\
DistMult \cite{Yang2014EmbeddingEA} & 87.1 & 86.2 & 86.7 \\
DistMult-HRS \cite{Zhang2018KnowledgeGE}  & \textbf{88.9} & 89.0 & 89.0 \\
AATE \cite{An2018AccurateTK} & 88.0 & 87.2 & 87.6 \\
ConvKB \cite{Nguyen2017ANE} & 87.6 & 88.8 & 88.2 \\
DOLORES \cite{Wang2018DOLORESDC} & 87.5 & 89.3 & 88.4 \\ \hline
Proposed method: LM-inspired KGE & 88.3 & \textbf{90.21} & \textbf{89.44} \\ \bottomrule 
\end{tabular}
}
\caption{Comparison of previous approaches with proposed method on triple classification task.}
\label{tab:results_2}   
\end{center}
\end{table}

\begin{table}[t]
\resizebox{\linewidth}{!}{%
\begin{tabular}{cccc}
\toprule
Methods & Strict F1 & Loose Micro & Loose Macro \\ \midrule
\citet{Ling2012FineGrainedER} & 52.30 & 69.30 & 69.90 \\
\citet{Ren2016LabelNR} & 49.44 & 68.75 & 68.75 \\
\citet{Ma2016LabelEF}. & 53.54 & 66.53 & 68.06 \\
\citet{Ren2016AFETAF} & 53.30 & 66.40 & 69.30 \\
\citet{Shimaoka2016AnAN} & \multirow{2}{*}{54.53} & \multirow{2}{*}{71.58} & \multirow{2}{*}{74.76} \\
(w/o Hand-Crafted features) &  &  &  \\
\citet{Shimaoka2016AnAN}  & \multirow{2}{*}{59.68} & \multirow{2}{*}{75.36} & \multirow{2}{*}{78.97} \\
(w/ Hand-Crafted features) &  & &  \\
\citet{Zhang2018FinegrainedET} & 60.05 & 75.52 & 78.67 \\
{Proposed Method} & \multirow{2}{*}{\textbf{61.10}} & \multirow{2}{*}{\textbf{75.70}} &  \multirow{2}{*}{\textbf{78.95}} \\
{(w/o Hand-Crafted features)} &  & & \\
{Proposed Method } & \multirow{2}{*}{\textbf{62.16}} & \multirow{2}{*}{\textbf{76.12}} & \multirow{2}{*}{\textbf{79.69}}  \\
{(w/ Hand-Crafted features)} &  & &  \\\bottomrule
\end{tabular}%
}
\caption{The performance of the proposed fine grain entity architecture to previous approaches on FIGER.}
\vspace{-1em}
\label{tab:figer}
\end{table}

We evaluate fine grain entity type prediction on the well-studied OntoNotes \cite{Hovy2006OntoNotesT9} and FIGER  \cite{Ling2012FineGrainedER} datasets. The OntoNotes dataset used here is a manually curated dataset by \citet{Gillick2014ContextDependentFE}, consisting of 89 different entity types. FIGER consists of 113 entity types, occuring in sentences from 780k Wikipedia articles and 434 news reports. We evaluate the joint KGE and Entity Typing model on WikiAuto and WikiMan, both introduced by \citet{Xin2018ImprovingNF}. WikiAuto is curated by distant supervision, with Freebase entities and types and sentence descriptions from Wikipedia articles. WikiMan is a manually curated dataset from Wikipedia articles with Freebase entities. 

Lastly, we evaluate the joint KGE and LM on WikiFact ~\cite{Ahn2017ANK}, built using the facts from Freebase and Wikipedia descriptions. The content of the dataset is limited to \textit{Film/Actor/} from Freebase. Further the anchor fact defined in the text of the dataset are not used for training the joint model. The description of the entities in the original dataset contain both the summary and the body from Wikipedia. The current study is performed by using the description from the summary section defined in the dataset. The joint model is trained and evaluated with the split of 80/10/10 for train, validation and test sets, respectively. 

\paragraph{Metrics}
KGE triple classification is evaluated through accuracy. The entity type model's performance is evaluated based on three common entity typing metrics---Strict F1, Loose Macro F1 and Loose Micro F1~\cite{Ling2012FineGrainedER}---while language modeling is measured by perplexity.

\paragraph{Previous Work as Baselines}
When possible, we directly compare our model's performance to that of previously published work.

\begin{table}[t]
\resizebox{\linewidth}{!}{%}
\centering{
\begin{tabular}{cccc}
\hline
Methods & Strict F1 & Macro F1 & Micro F1 \\ \hline
AFET \cite{Ren2016AFETAF} & 20.32 & 54.51 & 52.61 \\
KB only \cite{Xin2018ImprovingNF} & 35.12 & 70.49 & 63.36 \\
HNM \cite{Dong2015AHN}& 34.88 & 64.37 & 68.39 \\
SA \cite{Shimaoka2016AnAN}  & 42.77 & 72.40 & 74.91 \\
MA (KNET) \cite{Xin2018ImprovingNF} & 41.58 & 72.66 & 75.72 \\
KA (KNET) \cite{Xin2018ImprovingNF} & 45.49 & 72.46 & \textbf{76.22} \\
{Joint Model-Proposed} & \textbf{{46.18}} & {\textbf{72.78}} & {76.02} \\ \bottomrule
\end{tabular}
}
}
\caption{We compare previous techniques on the WIKIAUTO dataset for fine-grain typing. The proposed method outperforms all previous, comparable techniques. While techniques that utilize disambiguation to improve the results on the knowledge attention (e.g., KA + D (KNET) from \protect\citet{Xin2018ImprovingNF}) can yield very modest improvements, e.g., to 77 micro F1, due to the extra information used, those results are not directly comparable to the proposed model.}
\label{tab:wikiauto}
\end{table}

\begin{table}[t]
\resizebox{\linewidth}{!}{%
\centering
{
\begin{tabular}{cccc}
\hline
Methods & Strict F1 & Macro F1 & Micro F1 \\ \hline
AFET \cite{Ren2016AFETAF} & 18.00 & 56.33 & 56.52 \\
KB only \cite{Xin2018ImprovingNF} & 17.00 & 63.00 & 40.52 \\
HNM \cite{Dong2015AHN} & 15.00 & 64.75 & 65.30 \\
SA  \cite{Shimaoka2016AnAN} & 18.00 & 69.44 & 70.14 \\
MA (KNET) \cite{Xin2018ImprovingNF} & \textbf{26.00} & 71.19 & 72.08 \\
KA (KNET) \cite{Xin2018ImprovingNF} & 23.00 & 71.10 & 71.67 \\
{Joint Model- Proposed} & {{25.00}} & {\textbf{73.40}} & {\textbf{74.43}} \\ \bottomrule
\end{tabular}
}
}
\caption{We compare previous techniques on Wiki-MAN dataset for fine-grain entity type classification.}
\label{tab:wiki_man}
%\vspace{-1.1em}
\end{table}

\section{Results and Discussion}
\label{sec:results}
This section presents the results of our basic KGE, entity typing models, and the joint learning architecture and their comparison to previous methods. The models were trained using either a 16GB V100 or 11GB 2080 TI GPU (single GPU training only).

\subsection{The Effectiveness of a LM-inspired KGE}

The proposed knowledge graph embedding architecture (\cref{sec:kge-as-lm}) is trained for triple classification task: given an input triple $x_i$, predict whether the fact it represents is true or not. \Cref{tab:results_2} provides an overview of performance of our architecture in comparison to previously studies approaches, obtained from the corresponding paper.

Examining the results on WN11 and FB13, we see that in all but one case our approach improves upon the state of the art performance on triple classification task; in that one case (DistMult-HRS on WN11) our model was very competitive. These strong results support our hypothesis that language modeling principles can be an effective knowledge graph embedding technique. %
In examining per-relation performance on both WN11 and FB13, we observed an increase in the lower bound of accuracy results for relationships on both WordNet and Freebase, compared to  \citet{Socher2013ReasoningWN}. We see a rise in accuracy from \newcite{Socher2013ReasoningWN}'s 75.5\% to 81\% for the \textit{(domain region)} relation from WordNet. On Freebase, we see performance for the $institution$ relation goes from 77.2\% to 80.9\% with the current architecture.

Recently, \citet{Yao2019KGBERTBF} presented KG-BERT, which uses a pretrained BERT model to encode and classify triples. While this approach is empirically powerful, and surpasses our approach, we note that due to the limited training context of the current architecture, directly comparing those triple classification results with ours would be mischaracterizing the strengths and limitations of both approaches. Considering the training complexity and costs of transformer networks, our model presents an appealing balance between efficacy and efficiency. %

\subsection{The Effectiveness of Entity Typing with KGE-Inspired Models}
Our novel neural fine grain entity type prediction techniques is compared with previous approaches in \cref{tab:figer}. The neural architecture provides an improvement on FIGER in F1. To have a direct comparison, whe datasets used for the experiments are same as used by \citet{Shimaoka2016AnAN} and \citet{Zhang2018FinegrainedET}. Our method uses a margin based loss function to learn entity types, and outperforms all the previous methods \cite{Abhishek2017FineGrainedET,Ren2016AFETAF,Ren2016LabelNR} that learn fine grain entity type prediction through margin base loss functions and evaluated on the same datasets.

\begin{table*}
\centering
\resizebox{\textwidth}{!}{%
\begin{tabular}{@{}lllllllllll@{}}
\toprule
Dataset KGE & Dataset FNER & Strict F1 & Macro F1 & Micro -F1 & Accuracy & AUROC & AUCPR & F1 & Precision & Recall \\ \midrule
FB13 (our baseline) &  &  - &  - &  - & 90.21 & 0.96 & 0.95 & 0.9 & 0.89 & 0.91 \\
WN11 (our baseline)  &  &  - &  - &  - & 88.3 & 0.94 & 0.93 & 0.88 & 0.85 & 0.91 \\
FB15K (our baseline) &  &  - &  - &  - & 94.73 & 0.98 & 0.97 & 0.94 & 0.92 & 0.97 \\
 & OntoNotes (our baseline) & 53.22 & 69.36 & 61.65 &  - &  - &  - &  - &  - &  - \\
\midrule
%  \midrule
FB15k & OntoNotes & 53.33 & 70.47 & 62.95 & 93.43 & 0.97 & 0.97 & 0.93 & 0.9 & 0.97 \\
WN11 & OntoNotes & 52.79 & 69.12 & 61.62 & 87.61 & 0.93 & 0.94 & 0.88 & 0.83 & 0.94 \\
FB13 & OntoNotes & 53.34 & 70.81 & 63.44 & 89.79 & 0.96 & 0.96 & 0.9 & 0.87 & 0.93 \\
\end{tabular}%
}
\caption{We show the changes in performance we observe when training joint fine-grain entity type prediction and triple classification models (bottom portion) vs.\@ single-objective models (top portion). Joint training can lead to improvements on both KGE and FNER.}
\label{tab:joint_fner}
\end{table*}

\subsection{The Effectiveness of Joint KGE and Entity Typing}

Building on the baseline models, the joint model (\cref{fig:KGE_FNER}) addresses the implicit constraint given in the knowledge graph. The architecture learns to correlate the mention entities with the entities present in the context to addresses the problems of ``context-entity separation'' and ``text knowledge separation,'' as defined by \citet{Xin2018ImprovingNF}. The joint architecture is evaluated on the WikiAuto and WikiMan datasets. The model is trained with combination of FB15K dataset and WikiAuto to learn the both the factual information along with the entity typing structure. 
\Cref{tab:wikiauto,tab:wiki_man} provide an overview of results from current method and it comparisons with the previous techniques. 

We trained and tested the joint model on a combination of datasets for KGE and FNER; see \Cref{tab:joint_fner}. The results show the complementary nature of learning fine-grain entity types and knowledge graph embedding jointly with steady performances on either task with respect to their baselines.

\subsection{The Effectiveness of Joint KGE and Language Modeling}

\begin{figure}[t]
    \centering
     \includegraphics[scale=1.0, width=\linewidth]{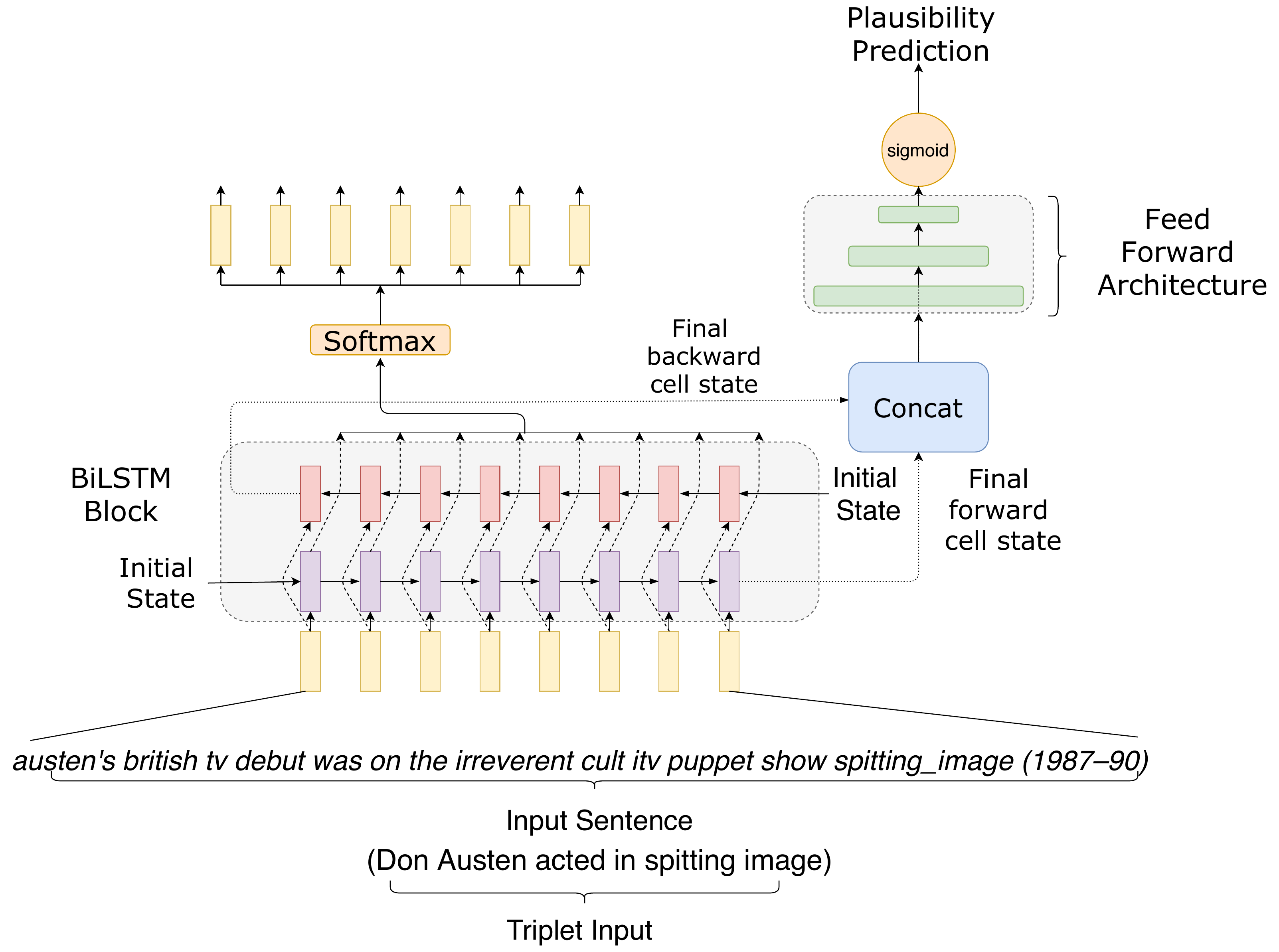}
    \caption{The architecture for joint learning of knowledge graph embedding with language model. We use an LSTM for the LM component, and a bi-LSTM for the KGE component. The LM LSTM and the forward portion of the bi-LSTM are the same, allowing the transfer of knowledge. The architecture takes in as input the whole sentence and the triplet to learn the semantic structure and factual information from the knowledge base.}
    \label{fig:kge_lm}
\end{figure}

\begin{table}[t]
\begin{subtable}[h]{.99\columnwidth}
\resizebox{.99\columnwidth}{!}{
\begin{tabular}{@{}lcc@{}}
\toprule
Model & Perplexity ($\downarrow$) & Acc ($\uparrow$) \\ \midrule
LSTM LM (baseline) & 440.72 & - \\
bi-LSTM KGE (baseline) & - & \textbf{94.22} \\
Joint LSTM LM + bi-LSTM KGE & \textbf{299.17} & 93.73 \\  \bottomrule
\end{tabular}
}
\caption{Performance of jointly learning an LSTM LM and bi-LSTM KGE.}
\label{tab:table_kge_lm:lstm:bilstm}
\end{subtable}
\hfill
\begin{subtable}{.99\columnwidth}
\resizebox{.99\columnwidth}{!}{
\begin{tabular}{@{}lcc@{}}
\toprule
Model & Perplexity ($\downarrow$) & Acc ($\uparrow$) \\ \midrule
LSTM LM (baseline) & 437.22 & - \\
LSTM KGE (baseline) & - & 90.66 \\
Joint LSTM LM + LSTM KGE & \textbf{353.72} & \textbf{93.6} \\ \bottomrule
\end{tabular}
}
\caption{Performance of jointly learning an LSTM LM and LSTM KGE.}
\label{tab:table_kge_lm:lstm:lstm}
\end{subtable}
\caption{We summarize the results from the joint KGE+LM experiments, learned from WikiFacts with a 70k word vocabulary. In \ref{tab:table_kge_lm:lstm:bilstm} we provide results for the architecture shown in \cref{fig:kge_lm} (a bi-LSTM KGE, whose forward cells are the cells of a unidirectional LSTM LM). %
In \ref{tab:table_kge_lm:lstm:lstm}, we provide results where we replace the bi-LSTM KGE with LSTM LM. %
\label{tab:table_kge_lm}
}
\end{table}

We examine the complementary nature of LM and KGE on the WikFacts dataset introduced by \citet{Ahn2017ANK}, which contains both sentences and KGE-style tuples. %
\Cref{fig:kge_lm} shows the architecture for jointly learning to embed a KG and model language. We use a single-layer LSTM (unidirectional: left-to-right) for language modeling, though the core KGE architecture relies on an bi-LSTM. We unify these by ensuring that the LM LSTM and the left-to-right portion of the KGE bi-LSTM use the same weights. %
We compare this joint approach to the same models trained separately and independently, without any weight sharing, evaluating the LMs on perplexity (lower is better) and KG prediction accuracy (higher is better). %
We use a vocabulary of the 70k most frequent words. %

As \cref{tab:table_kge_lm:lstm:bilstm} shows, while there is a very slight decrease in KG prediction accuracy, the distinct improvement in the performance of language model over the baseline LM demonstrates that joint learning is particularly effective for language modeling. %
This suggests that even simple joint learning can be an effective way of using stated knowledge to improve language modeling. %

\begin{table*}
\centering
\resizebox{\textwidth}{!}{%
\begin{tabular}{@{}ll@{}}
\toprule
 & \multicolumn{1}{c}{Sentences} \\ \midrule
Input sentence & \begin{tabular}[c]{@{}l@{}}stephen percy steve harris born 12 march 1956 is an english musician and songwriter known as the bassist occasional \\ keyboardist backing vocalist primary songwriter and founder of the british heavy metal band iron maiden \\ he is the only member of iron maiden to have remained in the band since their inception in 1975 \\ and along with guitarist dave murray to have appeared on all of their albums\end{tabular} \\ \midrule
Output (Joint model) & \begin{tabular}[c]{@{}l@{}}joseph john james unk born 5 april 1949 is an english musician and actor known as the greatest and guitarist \\ the vocalist guitarist songwriter and guitarist of  the band heavy metal band the band he is the founding child of the team \\ band have been by the band until its death in 2003 and toured with unk unk unk they have appeared in one of
\end{tabular} \\ \midrule
Output (baseline) & \begin{tabular}[c]{@{}l@{}}peter baron dickie unk born 11 august 1943 is an english singer and best and as the most and and and and  \\ lead songwriter and member of the heavy rock rock band unk side he is the third singer of the band band have been \\  with the band since its breakup in 1992 while cofounded with with dave tended has have collaborated on hundreds of their films
\end{tabular} \\ \bottomrule
\end{tabular}%
}
\caption{We provide an example of the sentence predicted by the language model jointly learned with knowledge graph embedding and the independently trained language model. %
Notice how some implicit constraints, learned from the KGE, are transferred to the language model.}
\label{tab:table_kge_sentence}
\end{table*}

While joint learning allowed the KG to help the LM, the reverse was not true. %
We speculate that this is in part because, from a language modeling perspective, the KGE model is able to consider both the forward and backward components. %
To test this, we replace the KGE bi-LSTM with the same unidirectional LSTM used by the LM. %
We show these results in \cref{tab:table_kge_lm:lstm:lstm}. %
Similar to the previous results, KGE allowed LM perplexity to decrease significantly. %
However, we also see that the LM yielded a 3 point absolute improvement in KG prediction, supporting our hypothesis. %

To further demonstrate how our joint learning method improves the semantic understanding of the language, we qualitatively examine the generative capacity of these LMs in \cref{tab:table_kge_sentence}. %
This provides an example of how joint training a KG and LM can improve output over a singly-trained LM on the same language data, and suggests that joint learning allows transfer of some implicit constraints in the language by learning the underlying relationships between the entities. %
While both are over-reliant on conjunctive structure, notice how the singly-trained baseline LM starts off alright, but then as the generation continues, loses coherence. Meanwhile, the jointly trained model maintains more coherence for longer. This suggests the KGE training is successfully transferring appropriate thematic/factive knowledge to the LM.

\section{Conclusion}
This work proposes a joint learning framework for learning real value representations of words, entities, and relations in a shared embedding space. Joint learning of factual representation with contextual understanding shows improvement in the learning of entity types. Learning the language model with knowledge graph embedding simultaneously enhances the performance on both modeling tasks. %
Our results suggest that language modeling could accelerate the study of schema-free approaches to both KGE and FNER, and strong performance can be obtained with comparatively simpler, resource-starved language models. %
This has promising implications for low-resource, and few-shot, and/or domain-specific information extraction needs.

{\small 
\paragraph{Acknowledgements}
We would like to thank members and affiliates of the UMBC CSEE Department, including Ankur Padia, Tim Finin, and Karuna Joshi. Some experiments were conducted on the UMBC HPCF. We'd also like to thank the reviewers for their comments and suggestions. %
This material is also based on research that is in part supported by the Air Force Research Laboratory (AFRL), DARPA, for the KAIROS program under agreement number FA8750-19-2-1003. The U.S.Government is authorized to reproduce and distribute reprints for Governmental purposes notwithstanding any copyright notation thereon. The views and conclusions contained herein are those of the authors and should not be interpreted as necessarily representing the official policies or endorsements, either express or implied, of the Air Force Research Laboratory (AFRL), DARPA, or the U.S. Government.
}

\bibliographystyle{acl_natbib}
\bibliography{references}

\begin{thebibliography}{59}
\expandafter\ifx\csname natexlab\endcsname\relax\def\natexlab#1{#1}\fi

\bibitem[{Abhishek et~al.(2017)Abhishek, Anand, and
  Awekar}]{Abhishek2017FineGrainedET}
Abhishek, Ashish Anand, and Amit Awekar. 2017.
\newblock Fine-grained entity type classification by jointly learning
  representations and label embeddings.
\newblock \emph{ArXiv}, abs/1702.06709.

\bibitem[{Ahn et~al.(2017)Ahn, Choi, P{\"a}rnamaa, and Bengio}]{Ahn2017ANK}
Sungjin Ahn, Heeyoul Choi, Tanel P{\"a}rnamaa, and Yoshua Bengio. 2017.
\newblock A neural knowledge language model.
\newblock \emph{ArXiv}, abs/1608.00318.

\bibitem[{An et~al.(2018)An, Chen, Han, and Sun}]{An2018AccurateTK}
Bo~An, Bo~Chen, Xianpei Han, and Le~Sun. 2018.
\newblock Accurate text-enhanced knowledge graph representation learning.
\newblock In \emph{NAACL-HLT}.

\bibitem[{Auer et~al.(2007)Auer, Bizer, Kobilarov, Lehmann, Cyganiak, and
  Ives}]{Auer2007DBpediaAN}
S{\"o}ren Auer, Christian Bizer, Georgi Kobilarov, Jens Lehmann, Richard
  Cyganiak, and Zachary~G. Ives. 2007.
\newblock Dbpedia: A nucleus for a web of open data.
\newblock In \emph{ISWC/ASWC}.

\bibitem[{Balazevic et~al.(2019)Balazevic, Allen, and
  Hospedales}]{balazevic2019tucker}
Ivana Balazevic, Carl Allen, and Timothy Hospedales. 2019.
\newblock Tucker: Tensor factorization for knowledge graph completion.
\newblock In \emph{EMNLP-IJCNLP}, pages 5188--5197.

\bibitem[{Bengio et~al.(2000)Bengio, Ducharme, Vincent, and
  Janvin}]{Bengio2000ANP}
Yoshua Bengio, R{\'e}jean Ducharme, Pascal Vincent, and Christian Janvin. 2000.
\newblock A neural probabilistic language model.
\newblock \emph{J. Mach. Learn. Res.}, 3:1137--1155.

\bibitem[{Bollacker et~al.(2008)Bollacker, Evans, Paritosh, Sturge, and
  Taylor}]{Bollacker2008FreebaseAC}
Kurt~D. Bollacker, C.~J. Evans, Praveen Paritosh, Tim Sturge, and Jamie Taylor.
  2008.
\newblock Freebase: a collaboratively created graph database for structuring
  human knowledge.
\newblock In \emph{SIGMOD Conference}.

\bibitem[{Bordes et~al.(2013)Bordes, Usunier, Garc{\'i}a-Dur{\'a}n, Weston, and
  Yakhnenko}]{Bordes2013TranslatingEF}
Antoine Bordes, Nicolas Usunier, Alberto Garc{\'i}a-Dur{\'a}n, Jason Weston,
  and Oksana Yakhnenko. 2013.
\newblock Translating embeddings for modeling multi-relational data.
\newblock In \emph{NIPS}.

\bibitem[{Bosselut et~al.(2019)Bosselut, Rashkin, Sap, Malaviya, Celikyilmaz,
  and Choi}]{bosselut-etal-2019-comet}
Antoine Bosselut, Hannah Rashkin, Maarten Sap, Chaitanya Malaviya, Asli
  Celikyilmaz, and Yejin Choi. 2019.
\newblock {COMET}: Commonsense transformers for automatic knowledge graph
  construction.
\newblock In \emph{Proceedings of the 57th Annual Meeting of the Association
  for Computational Linguistics}. Association for Computational Linguistics.

\bibitem[{Cochez et~al.(2017)Cochez, Ristoski, Ponzetto, and
  Paulheim}]{cochez2017global}
Michael Cochez, Petar Ristoski, Simone~Paolo Ponzetto, and Heiko Paulheim.
  2017.
\newblock Global rdf vector space embeddings.
\newblock In \emph{International Semantic Web Conference}, pages 190--207.
  Springer.

\bibitem[{Dettmers et~al.(2018)Dettmers, Minervini, Stenetorp, and
  Riedel}]{dettmers2017convolutional}
Tim Dettmers, Pasquale Minervini, Pontus Stenetorp, and Sebastian Riedel. 2018.
\newblock Convolutional 2d knowledge graph wmbeddings.
\newblock In \emph{AAAI}.

\bibitem[{Devlin et~al.(2019)Devlin, Chang, Lee, and
  Toutanova}]{devlin-etal-2019-bert}
Jacob Devlin, Ming-Wei Chang, Kenton Lee, and Kristina Toutanova. 2019.
\newblock {BERT}: Pre-training of deep bidirectional transformers for language
  understanding.
\newblock In \emph{NAACL}.

\bibitem[{Dong et~al.(2015)Dong, Wei, Sun, Zhou, and Xu}]{Dong2015AHN}
Li~Dong, Furu Wei, Hong Sun, Ming Zhou, and Ke~Xu. 2015.
\newblock A hybrid neural model for type classification of entity mentions.
\newblock In \emph{IJCAI}.

\bibitem[{Gillick et~al.(2014)Gillick, Lazic, Ganchev, Kirchner, and
  Huynh}]{Gillick2014ContextDependentFE}
Daniel Gillick, Nevena Lazic, Kuzman Ganchev, Jesse Kirchner, and David Huynh.
  2014.
\newblock Context-dependent fine-grained entity type tagging.
\newblock \emph{ArXiv}, abs/1412.1820.

\bibitem[{Hochreiter and Schmidhuber(1997)}]{Hochreiter1997LongSM}
Sepp Hochreiter and J{\"u}rgen Schmidhuber. 1997.
\newblock Long short-term memory.
\newblock \emph{Neural Computation}, 9:1735--1780.

\bibitem[{Hovy et~al.(2006)Hovy, Marcus, Palmer, Ramshaw, and
  Weischedel}]{Hovy2006OntoNotesT9}
Eduard~H. Hovy, Mitchell~P. Marcus, Martha Palmer, Lance~A. Ramshaw, and
  Ralph~M. Weischedel. 2006.
\newblock Ontonotes: The 90\% solution.
\newblock In \emph{HLT-NAACL}.

\bibitem[{Ioffe and Szegedy(2015)}]{Ioffe2015BatchNA}
Sergey Ioffe and Christian Szegedy. 2015.
\newblock Batch normalization: Accelerating deep network training by reducing
  internal covariate shift.
\newblock \emph{ArXiv}, abs/1502.03167.

\bibitem[{Ji et~al.(2015)Ji, He, Xu, Liu, and Zhao}]{Ji2015KnowledgeGE}
Guoliang Ji, Shizhu He, Liheng Xu, Kang Liu, and Jun Zhao. 2015.
\newblock Knowledge graph embedding via dynamic mapping matrix.
\newblock In \emph{ACL}.

\bibitem[{Ji et~al.(2016)Ji, Liu, He, and Zhao}]{Ji2016KnowledgeGC}
Guoliang Ji, Kang Liu, Shizhu He, and Jun Zhao. 2016.
\newblock Knowledge graph completion with adaptive sparse transfer matrix.
\newblock In \emph{AAAI}.

\bibitem[{Kingma and Ba(2014)}]{Kingma2014AdamAM}
Diederik~P. Kingma and Jimmy Ba. 2014.
\newblock Adam: A method for stochastic optimization.
\newblock \emph{CoRR}, abs/1412.6980.

\bibitem[{Krompa{\ss} et~al.(2013)Krompa{\ss}, Nickel, Jiang, and
  Tresp}]{krompass2013non}
Denis Krompa{\ss}, Maximilian Nickel, Xueyan Jiang, and Volker Tresp. 2013.
\newblock Non-negative tensor factorization with rescal.
\newblock In \emph{Tensor Methods for Machine Learning, ECML workshop}.

\bibitem[{Lin et~al.(2015)Lin, Liu, Sun, Liu, and Zhu}]{Lin2015LearningEA}
Yankai Lin, Zhiyuan Liu, Maosong Sun, Yang Liu, and Xuan Zhu. 2015.
\newblock Learning entity and relation embeddings for knowledge graph
  completion.
\newblock In \emph{AAAI}.

\bibitem[{Ling and Weld(2012)}]{Ling2012FineGrainedER}
Xiao Ling and Daniel~S. Weld. 2012.
\newblock Fine-grained entity recognition.
\newblock In \emph{AAAI}.

\bibitem[{{Logan IV} et~al.(2019){Logan IV}, Liu, Peters, Gardner, and
  Singh}]{RobertLLogan2019BaracksWH}
Robert~L. {Logan IV}, Nelson~F. Liu, Matthew~E. Peters, Matt Gardner, and
  Sameer Singh. 2019.
\newblock Barack's wife hillary: Using knowledge-graphs for fact-aware language
  modeling.
\newblock In \emph{ACL}.

\bibitem[{Ma et~al.(2016)Ma, Cambria, and Gao}]{Ma2016LabelEF}
Yukun Ma, Erik Cambria, and Sa~Gao. 2016.
\newblock Label embedding for zero-shot fine-grained named entity typing.
\newblock In \emph{COLING}.

\bibitem[{Mikolov et~al.(2010)Mikolov, Karafi{\'a}t, Burget, {\v C}ernock{\'y},
  and Khudanpur}]{Mikolov2010RecurrentNN}
Tomas Mikolov, Martin Karafi{\'a}t, Luk{\'a}s Burget, Jan {\v C}ernock{\'y},
  and Sanjeev Khudanpur. 2010.
\newblock Recurrent neural network based language model.
\newblock In \emph{INTERSPEECH}.

\bibitem[{Minervini et~al.(2017)Minervini, Costabello, Mu{\~n}oz, Nov{\'a}cek,
  and Vandenbussche}]{Minervini2017RegularizingKG}
Pasquale Minervini, Luca Costabello, Emir Mu{\~n}oz, V{\'i}t Nov{\'a}cek, and
  Pierre-Yves Vandenbussche. 2017.
\newblock Regularizing knowledge graph embeddings via equivalence and inversion
  axioms.
\newblock In \emph{ECML/PKDD}.

\bibitem[{Nair and Hinton(2010)}]{Nair2010RectifiedLU}
Vinod Nair and Geoffrey~E. Hinton. 2010.
\newblock Rectified linear units improve restricted boltzmann machines.
\newblock In \emph{ICML}.

\bibitem[{Nguyen et~al.(2017)Nguyen, Nguyen, Nguyen, and Phung}]{Nguyen2017ANE}
Dai~Quoc Nguyen, Tu~Dinh Nguyen, Dat~Quoc Nguyen, and Dinh~Q. Phung. 2017.
\newblock A novel embedding model for knowledge base completion based on
  convolutional neural network.
\newblock In \emph{NAACL-HLT}.

\bibitem[{Nickel et~al.(2011)Nickel, Tresp, and Kriegel}]{Nickel2011ATM}
Maximilian Nickel, Volker Tresp, and Hans-Peter Kriegel. 2011.
\newblock A three-way model for collective learning on multi-relational data.
\newblock In \emph{ICML}.

\bibitem[{Padia et~al.(2019)Padia, Kalpakis, Ferraro, and
  Finin}]{Padia2019KnowledgeGF}
Ankur Padia, Konstantinos Kalpakis, Francis Ferraro, and Timothy~W. Finin.
  2019.
\newblock Knowledge graph fact prediction via knowledge-enriched tensor
  factorization.
\newblock \emph{J. Web Semant.}, 59.

\bibitem[{Pennington et~al.(2014)Pennington, Socher, and
  Manning}]{Pennington2014GloveGV}
Jeffrey Pennington, Richard Socher, and Christopher~D. Manning. 2014.
\newblock Glove: Global vectors for word representation.
\newblock In \emph{EMNLP}.

\bibitem[{Peters et~al.(2018)Peters, Neumann, Iyyer, Gardner, Clark, Lee, and
  Zettlemoyer}]{peters2018elmo}
Matthew Peters, Mark Neumann, Mohit Iyyer, Matt Gardner, Christopher Clark,
  Kenton Lee, and Luke Zettlemoyer. 2018.
\newblock Deep contextualized word representations.
\newblock In \emph{NAACL}.

\bibitem[{Peters et~al.(2019)Peters, Neumann, Logan, Schwartz, Joshi, Singh,
  and Smith}]{peters-etal-2019-knowledge}
Matthew~E. Peters, Mark Neumann, Robert Logan, Roy Schwartz, Vidur Joshi,
  Sameer Singh, and Noah~A. Smith. 2019.
\newblock Knowledge enhanced contextual word representations.
\newblock In \emph{EMNLP-IJCNLP}.

\bibitem[{Petroni et~al.(2019)Petroni, Rockt{\"a}schel, Riedel, Lewis, Bakhtin,
  Wu, and Miller}]{petroni-etal-2019-language}
Fabio Petroni, Tim Rockt{\"a}schel, Sebastian Riedel, Patrick Lewis, Anton
  Bakhtin, Yuxiang Wu, and Alexander Miller. 2019.
\newblock Language models as knowledge bases?
\newblock In \emph{EMNLP-IJCNLP}.

\bibitem[{Ren et~al.(2016{\natexlab{a}})Ren, He, Qu, Huang, Ji, and
  Han}]{Ren2016AFETAF}
Xiang Ren, Wenqi He, Meng Qu, Lifu Huang, Heng Ji, and Jiawei Han.
  2016{\natexlab{a}}.
\newblock Afet: Automatic fine-grained entity typing by hierarchical
  partial-label embedding.
\newblock In \emph{EMNLP}.

\bibitem[{Ren et~al.(2016{\natexlab{b}})Ren, He, Qu, Voss, Ji, and
  Han}]{Ren2016LabelNR}
Xiang Ren, Wenqi He, Meng Qu, Clare~R. Voss, Heng Ji, and Jiawei Han.
  2016{\natexlab{b}}.
\newblock Label noise reduction in entity typing by heterogeneous partial-label
  embedding.
\newblock \emph{ArXiv}, abs/1602.05307.

\bibitem[{Ristoski and Paulheim(2016)}]{ristoski2016rdf2vec}
Petar Ristoski and Heiko Paulheim. 2016.
\newblock {RDF}2{V}ec: {RDF} graph embeddings for data mining.
\newblock In \emph{International Semantic Web Conference}, pages 498--514.
  Springer.

\bibitem[{Rosenfeld(1994)}]{rosenfeld1994phd}
Ronald Rosenfeld. 1994.
\newblock \emph{Adaptive statistical language modeling: A maximum
  entropyapproach}.
\newblock Ph.D. thesis, Computer Science Department, Carnegie Mellon
  University.

\bibitem[{Schuster and Paliwal(1997)}]{Schuster1997BidirectionalRN}
Mike Schuster and Kuldip~K. Paliwal. 1997.
\newblock Bidirectional recurrent neural networks.
\newblock \emph{IEEE Trans. Signal Processing}, 45:2673--2681.

\bibitem[{Shimaoka et~al.(2016)Shimaoka, Stenetorp, Inui, and
  Riedel}]{Shimaoka2016AnAN}
Sonse Shimaoka, Pontus Stenetorp, Kentaro Inui, and Sebastian Riedel. 2016.
\newblock An attentive neural architecture for fine-grained entity type
  classification.
\newblock In \emph{AKBC@NAACL-HLT}.

\bibitem[{Socher et~al.(2013)Socher, Chen, Manning, and
  Ng}]{Socher2013ReasoningWN}
Richard Socher, Danqi Chen, Christopher~D. Manning, and Andrew~Y. Ng. 2013.
\newblock Reasoning with neural tensor networks for knowledge base completion.
\newblock In \emph{NIPS}.

\bibitem[{Strapparava and Valitutti(2004)}]{Strapparava2004WordNetAA}
Carlo Strapparava and Alessandro Valitutti. 2004.
\newblock Wordnet affect: an affective extension of wordnet.
\newblock In \emph{LREC}.

\bibitem[{Suchanek et~al.(2007)Suchanek, Kasneci, and
  Weikum}]{Suchanek2007YagoAC}
Fabian~M. Suchanek, Gjergji Kasneci, and Gerhard Weikum. 2007.
\newblock Yago: a core of semantic knowledge.
\newblock In \emph{WWW '07}.

\bibitem[{Toutanova and Chen(2015)}]{toutanova2015observed}
Kristina Toutanova and Danqi Chen. 2015.
\newblock Observed versus latent features for knowledge base and text
  inference.
\newblock In \emph{Proceedings of the 3rd Workshop on Continuous Vector Space
  Models and their Compositionality}, pages 57--66.

\bibitem[{Wang et~al.(2018)Wang, Kulkarni, and Wang}]{Wang2018DOLORESDC}
Haoyu Wang, Vivek Kulkarni, and William~Yang Wang. 2018.
\newblock Dolores: Deep contextualized knowledge graph embeddings.
\newblock \emph{ArXiv}, abs/1811.00147.

\bibitem[{Wang et~al.(2014)Wang, Zhang, Feng, and Chen}]{Wang2014KnowledgeGE}
Zhen Wang, Jianwen Zhang, Jianlin Feng, and Zhigang Chen. 2014.
\newblock Knowledge graph embedding by translating on hyperplanes.
\newblock In \emph{AAAI}.

\bibitem[{Wang and Li(2016)}]{Wang2016TextEnhancedRL}
Zhigang Wang and Juan-Zi Li. 2016.
\newblock Text-enhanced representation learning for knowledge graph.
\newblock In \emph{IJCAI}.

\bibitem[{Weston et~al.(2013)Weston, Bordes, Yakhnenko, and
  Usunier}]{Weston2013ConnectingLA}
Jason Weston, Antoine Bordes, Oksana Yakhnenko, and Nicolas Usunier. 2013.
\newblock Connecting language and knowledge bases with embedding models for
  relation extraction.
\newblock In \emph{EMNLP}.

\bibitem[{Xiao et~al.(2016)Xiao, Huang, and Zhu}]{Xiao2016TransGA}
Han Xiao, Minlie Huang, and Xiaoyan Zhu. 2016.
\newblock Transg : A generative model for knowledge graph embedding.
\newblock In \emph{ACL}.

\bibitem[{Xin et~al.(2018{\natexlab{a}})Xin, Lin, Liu, and
  Sun}]{Xin2018ImprovingNF}
Ji~Xin, Yankai Lin, Zhiyuan Liu, and Maosong Sun. 2018{\natexlab{a}}.
\newblock Improving neural fine-grained entity typing with knowledge attention.
\newblock In \emph{AAAI}.

\bibitem[{Xin et~al.(2018{\natexlab{b}})Xin, Zhu, Han, Liu, and
  Sun}]{Xin2018PutIB}
Ji~Xin, Hao Zhu, Xu~Han, Zhiyuan Liu, and Maosong Sun. 2018{\natexlab{b}}.
\newblock Put it back: Entity typing with language model enhancement.
\newblock In \emph{EMNLP}.

\bibitem[{Xu and Barbosa(2018)}]{Xu2018NeuralFE}
Peng Xu and Denilson Barbosa. 2018.
\newblock Neural fine-grained entity type classification with hierarchy-aware
  loss.
\newblock In \emph{NAACL-HLT}.

\bibitem[{Yang et~al.(2014)Yang, tau Yih, He, Gao, and
  Deng}]{Yang2014EmbeddingEA}
Bishan Yang, Wen tau Yih, Xiaodong He, Jianfeng Gao, and Li~Deng. 2014.
\newblock Embedding entities and relations for learning and inference in
  knowledge bases.
\newblock \emph{CoRR}, abs/1412.6575.

\bibitem[{Yang et~al.(2019)Yang, Dai, Yang, Carbonell, Salakhutdinov, and
  Le}]{yang2019xlnet}
Zhilin Yang, Zihang Dai, Yiming Yang, Jaime Carbonell, Russ~R Salakhutdinov,
  and Quoc~V Le. 2019.
\newblock Xlnet: Generalized autoregressive pretraining for language
  understanding.
\newblock In \emph{NeurIPS}.

\bibitem[{Yao et~al.(2019)Yao, Mao, and Luo}]{Yao2019KGBERTBF}
Liang Yao, Chengsheng Mao, and Yuan Luo. 2019.
\newblock Kg-bert: Bert for knowledge graph completion.
\newblock \emph{ArXiv}, abs/1909.03193.

\bibitem[{Zhang et~al.(2018{\natexlab{a}})Zhang, Duh, and
  Durme}]{Zhang2018FinegrainedET}
Sheng Zhang, Kevin Duh, and Benjamin~Van Durme. 2018{\natexlab{a}}.
\newblock Fine-grained entity typing through increased discourse context and
  adaptive classification thresholds.
\newblock In \emph{*SEM@NAACL-HLT}.

\bibitem[{Zhang et~al.(2018{\natexlab{b}})Zhang, Zhuang, Qu, Lin, and
  He}]{Zhang2018KnowledgeGE}
Zhao Zhang, Fuzhen Zhuang, Meng Qu, Fen Lin, and Qing He. 2018{\natexlab{b}}.
\newblock Knowledge graph embedding with hierarchical relation structure.
\newblock In \emph{EMNLP}.

\bibitem[{Zhang et~al.(2019)Zhang, Han, Liu, Jiang, Sun, and
  Liu}]{zhang-etal-2019-ernie}
Zhengyan Zhang, Xu~Han, Zhiyuan Liu, Xin Jiang, Maosong Sun, and Qun Liu. 2019.
\newblock {ERNIE}: Enhanced language representation with informative entities.
\newblock In \emph{ACL}.

\end{thebibliography}

\end{document}